\documentclass[10pt, conference, compsocconf]{IEEEtran}
\ifCLASSINFOpdf
\else
\fi

\usepackage{amssymb}
\usepackage[cmex10]{amsmath}
\usepackage{graphicx}
\usepackage{subfigure}
 \usepackage{multirow}
\usepackage{balance} 
\usepackage[vlined,boxed,commentsnumbered, ruled, linesnumbered]{algorithm2e}
\usepackage{amsfonts,amssymb}
\usepackage{setspace}
\usepackage{amsmath}
\usepackage{mathrsfs}
\usepackage{caption}
\usepackage{subfigure}
\usepackage{multirow}
\usepackage{amssymb}

\begin{document}
%
\title{Defending Adversarial Attacks via Semantic Feature Manipulation}

\author{
\IEEEauthorblockN{Shuo Wang}
\IEEEauthorblockA{CSIRO $\&$   Monash University  \\
Melbourne, Australia\\
shuo.wang@csiro.au/monash.edu}
\and
\IEEEauthorblockN{Tianle Chen}
\IEEEauthorblockA{Monash University\\
Melbourne, Australia\\
tianlec@student.monash.edu}
\and
\IEEEauthorblockN{Shangyu Chen}
\IEEEauthorblockA{University of Melbourne\\
Melbourne, Australia\\
shangyuc@student.unimelb.edu.au }
%
\and
\IEEEauthorblockN{Surya Nepal}
\IEEEauthorblockA{CSIRO\\
Melbourne, Australia\\
surya.nepal@data61.csiro.au}
\and
\IEEEauthorblockN{Marthie Grobler}
\IEEEauthorblockA{CSIRO\\
Melbourne, Australia\\
Marthie.Grobler@data61.csiro.au}
\and
\IEEEauthorblockN{Carsten Rudolph}
\IEEEauthorblockA{Monash University\\
Melbourne, Australia\\
carsten.rudolph@monash.edu}
}


%


\maketitle

\begin{abstract}
Machine learning models have demonstrated vulnerability to adversarial attacks, more specifically misclassification of adversarial examples. 
In this paper, we propose a one-off and attack-agnostic Feature Manipulation (FM)-Defense to detect and purify adversarial examples in an interpretable and efficient manner. 
The intuition is that the classification result of a normal image is generally resistant to non-significant intrinsic feature changes, e.g., varying thickness of handwritten digits. In contrast, adversarial examples are sensitive to such changes since the perturbation lacks transferability. 
To enable manipulation of features, a combo-variational autoencoder is applied to learn disentangled latent codes that reveal semantic features. 
The resistance to classification change over the morphs, derived by varying and reconstructing latent codes, is used to detect suspicious inputs. 
Further, combo-VAE is enhanced to purify the adversarial examples with good quality by considering both class-shared and class-unique features. We empirically demonstrate the effectiveness of detection and the quality of purified instance. 
Our experiments on three datasets show that FM-Defense can detect nearly $100\%$ of adversarial examples produced by different state-of-the-art adversarial attacks. It achieves more than $99\%$ overall purification accuracy on the suspicious instances that close the manifold of normal examples.
\end{abstract}

\section{Introduction}
The existence of adversarial examples causes serious security concerns, particularly in casting doubt on the reliability of Deep Neural Networks (DNNs) in the case of image classification.
These adversarial examples can be generated by adding visually imperceptible perturbations into a normal image to cause a DNN to mislabel the perturbed images with high confidence \cite{papernot2016limitations,liu2016delving}. Such adversarial attacks may lead to catastrophic consequences in applications such as disease diagnosis and self-driving cars. 
Existing defensive approaches proposed in the literature to defeat adversarial threats can be mainly categorized as adversarial training, defensive distillation and detecting/purifying adversarial examples. The first two methods involve the modification of the protected classifier or require knowledge of the process to generate adversarial examples. The third method aims at identifying suspicious inputs from normal inputs using hand-crafted statistical features \cite{grosse2017statistical}, separate classification networks \cite{metzen2017detecting,tian2018detecting} or autoencoders \cite{meng2017magnet}. 
Unfortunately, the Carlini-Wagner (CW) attack \cite{carlini2017adversarial} has demonstrated that most existing detection approaches can be evaded. The efficient detection of adversary examples without knowledge of adversarial example generation, therefore, remains a challenge for machine learning and security communities. This work aims at an efficient detecting and purifying defense.


Existing detection-based defensive approaches include Defense-GAN \cite{samangouei2018defense}, MagNet \cite{meng2017magnet}, FBGAN \cite{bao2018featurized} and Image Transformation-based detection \cite{tian2018detecting}. 
Defense-GAN trains a GAN to generate the manifold of unperturbed images and then finds the nearest point on the manifold to the adversarial example as the denoising result. 
MagNet applies detector networks to learn and differentiate between normal and adversarial examples by approximating the manifold of normal examples. Applying the reformer network moves the adversarial examples towards the manifold of normal examples to correctly reconstruct adversarial examples with small perturbation. 
FBGAN extracts the semantic features of the input images and reconstructs the denoised images from these features. It uses the generative capability of Bidirectional GAN and the mutual information (MI) regularization between all the latent codes and the generated images for disentanglement. 
Image Transformation-based detection applies certain transformation operations on an image to generate several transformed images. Then the classification results of these transformed images are used to distinguish between the normal and the adversarial. 
These approaches present two significant drawbacks: 

\begin{figure*}[!htb]
\centering
\setlength{\abovecaptionskip}{-0.05cm}
\setlength{\belowcaptionskip}{-0.2cm}
\subfigure[Framework of our defense]{
\includegraphics[width=6.5in,height=3.1in]{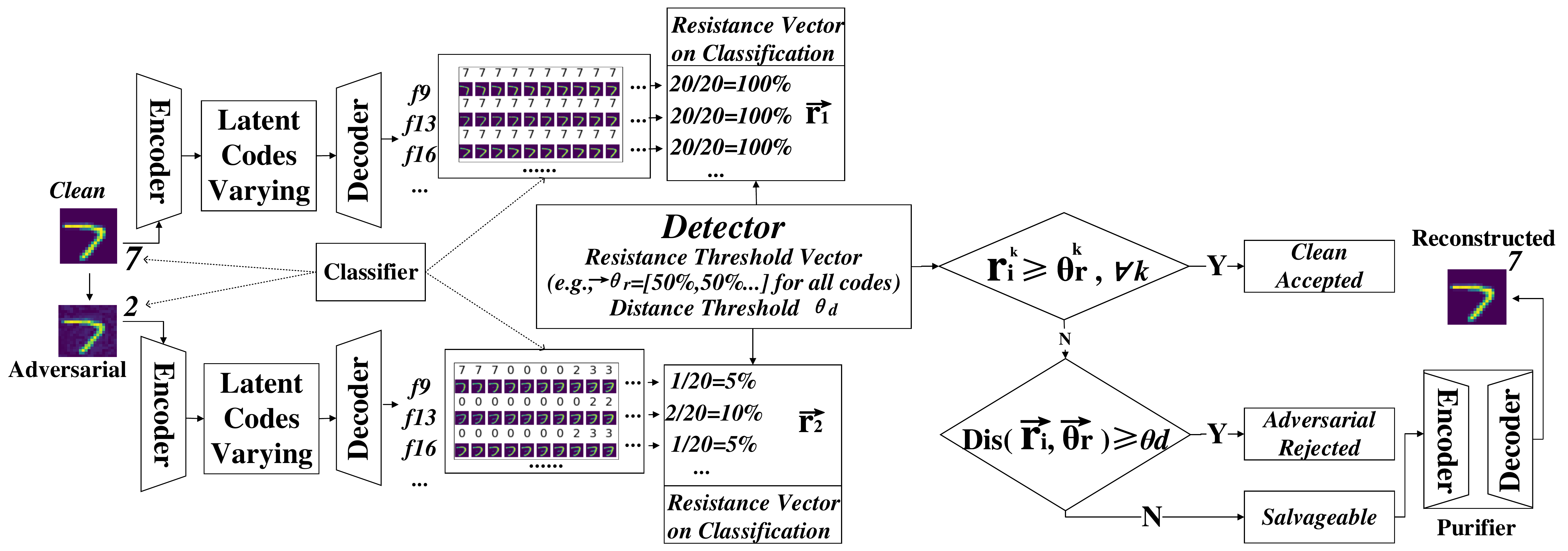}  

}
\subfigure[Illustration of our defense]{    \includegraphics[width=6.5in,height=3.2in]{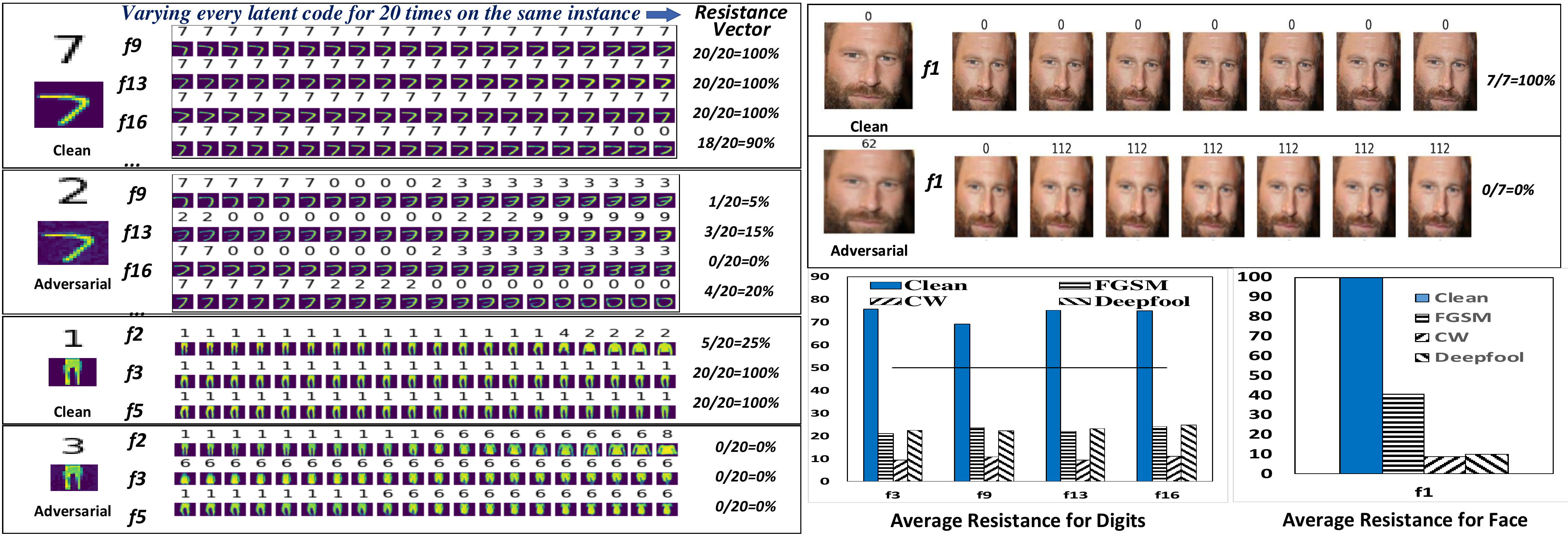}     }

\caption{Illustration of the feature manipulation on normal and adversarial images  from the MNIST/FMNIST/CelebA dataset. \small{The first sub-figures are the illustration of the FM-Defense by feature manipulating. We select ten latent codes that reveal ten semantic features, e.g., the thickness (f13), azimuth (f16) and stretching (f9) for MNIST digit "7", style features (f2, 3, 5) for trouser shape in FMNIST, and face expression feature f1 for CelebA face images. For each instance, we change every latent code for $n$ times to get $n$ morphs, and record the predicted label by a pre-trained classifier. The classification accuracy resistance is evaluated by the proportion of unchanged predictions among morphs. The histogram is the average resistance on classification accuracy for 2000 clean and perturbed instances derived from four adversarial attacks on MNIST and CelebA, respectively.} }
\end{figure*}

{\it Feasibility.} 
Element-wise metrics, such as the pixel-wise squared error, are commonly adopted for reconstruction error based adversarial detection, e.g., Magnet. As reconstruction error is a continuous value, a threshold can be set as a hyperparameter to decide whether the input is normal or adversarial. However, the reconstruction error of the perturbed images derived from oblivious attacks (such as the CW attack) is very likely similar to normal images. The threshold should be as low as possible to identify slightly perturbed adversarial examples, considering that too low would significantly misjudge normal examples.
Besides, for image transformation-based detection, a mixture of normal and adversarial examples is required in the training set to train the classifier. This causes the high computational cost to generate adversarial examples and adversarial classifiers, particularly for more complex datasets and stronger attacks. Further, the impact of external features transformation, such as image rotation and shifting, is not consistent and general for different instances. Besides, the background of an image adds a large number of extra features to the object, which is also sensitive to the external transformation. 


{\it Completeness.} Generally, there are two types of semantic features: class-shared (such as the thickness of the handwritten digits, the facial expression of the face images) and class-unique (such as different handwriting style for each digit or the identification of face). However, the reformer of MagNet and the generator of FBGAN and Defense-GAN can only reconstruct/generate the purified instance using some commonly shared feature. It leads to loss of class-unique features, for example, different writing styles exclusive to the digit "2" (such as a flat stroke or across loop bottom); such features become inactive for some classification tasks, e.g., writer identification or face recognition. As the high dimension space input, e.g., image, always lies in a complex manifold, the underlying data distribution could be very complex. The purifier is used to build complex enough models to capture the true posterior by utilizing both class-shared and class-unique features.

In this paper, we propose an adversarial example detection and purification method, named Feature-Manipulation defense (FM-Defense), to address these two concerns. It can effectively defeat the state-of-the-art adversarial attacks, including CW attack.
The intuition is that the classification result of a normal image is generally resistant to non-significant intrinsic feature changes, e.g., varying thickness of handwritten digits or the facial expression. Namely, the classification results of a normal digit and its morphs, derived by varying thickness, are very likely to be stable, since the significant features are retained. In contrast, adversarial examples are sensitive to such changes. The reason is that the unstructured perturbation is designed for a single image, and may cause various impacts on its morphs due to lack of transferability. 
Figure 1 demonstrates our intuition using a handwritten digit, trouser shape and face image.

The key point to implement our intuition is how to manipulate the intrinsic feature. Consequently, a one-off combo-variational autoencoder (combo-VAE) is applied to learn disentangled low-dimensional latent codes, i.e., one latent code only affects one semantic feature. The learned latent codes are disentangled, easy to control, and composed of abundant internal semantic features, instead of external features such as image rotation and shifting. 
The resistance of classification accuracy when manipulating the intrinsic features via disentangled latent codes is used to detect suspicious inputs. As demonstrated in the histograms of Figure 1, the classification accuracy change of clean instances is more consistent than that of adversarial instances. Hence, a simple threshold of classification accuracy resistance can be set to easily distinguish normal and adversarial images instead of training the adversarial classifier. 
Further, the combo-VAE is also applied to purify the suspicious instances close to the manifold of normal examples by reconstructing using both class-shared and class-unique features to move them towards the manifold.
To the best of our knowledge, the FM-Defense is the first attempt to apply disentangled learning for effective defense against oblivious adversarial attacks via both detection and purification, with good interpretability, feasibility, and completeness.

Our contributions are summarized as follows. We first present a key intuition that adversarial examples are generally more sensitive to intrinsic feature changes than normal images. Based on this intuition, we then propose feature manipulation-based adversarial example detection and purification method, FM-Defense. We use a combo-VAE to manipulate the feature in an easy and interpretable manner. 
 Besides detection, the combo-VAE is used to purify the suspicious inputs by reconstructing images based on both class-unique and class-shared components. 
It can improve the completeness of the reconstructed instance for purification.
We implement and evaluate the FM-Defense on three image datasets, MNIST, FMNIST and CelebA, which shows the superior performance in defending against various adversarial attacks.
\section{Background and related work}
\subsection{ Autoencoders and $\beta$-VAE}
Autoencoders (AEs) are common deep models in unsupervised learning \cite{bengio2013representation}. They aim to represent high-dimensional data through the low-dimensional latent layer, a.k.a. bottleneck vector or code. Architecturally, AEs consist of two parts, the encoder and decoder. The encoder part takes the input $x \in R^d$ and maps it to $z$ (the latent variable of the bottleneck vector). The decoder tries to reconstruct the input data from $z$. The training process of autoencoders is to minimize the reconstruction error. Formally, we can define the encoder and the decoder as transitions $ \tau_1$ and $ \tau_2$:
\begin{equation}
\begin{aligned}
\tau_1(X)\rightarrow Z \\
\tau_2(Z)\rightarrow \hat{X} \\
\tau_1,\tau_2=\underset{\tau_1,\tau_2}{argmin}\left \| X-\hat{X} \right \|^2
\end{aligned}
\end{equation}
The VAEs model shares the same structure with the autoencoders, but is based on an assumption that the latent variables follow some kind of distribution, such as Gaussian or uniform distribution. It uses variational inference for the learning of the latent variables. In VAEs the hypothesis is that the data is generated by a directed graphical model $p(x|z)$ and the encoder is to learn an approximation $q_{\phi} (z|x)$ to the posterior distribution $p_{\theta}(z|x)$. The VAE optimizes the variational lower bound:
\begin{equation}
L(\theta ,\phi ;x) = KL(q_{\phi }(z|x)||p_{\theta }(z)) - \mathbf{E}_{q_{\phi }(z|x)}[log p_{\theta }(x|z)] 
\end{equation}
The left part is the regularization term to match the posterior of $z$ conditional on $x$, i.e., $ q_{\phi }(z|x)$, to a target distribution $ p_{\theta }(z)$ by the KL divergence. The right part denotes the reconstruction loss for a specific sample $x$. In a training batch, the loss can be averaged as:
\begin{equation}
\begin{aligned}
L_{VAE} = \mathbf{E}_{p_{data}(x)}[L(\theta ,\phi ;x) ] \\
=\mathbf{E}_{p_{data}(x)}[KL(q_{\phi }(z|x)||p_{\theta }(z)) ] -\\
\textbf{E}_{p_{data}(x)}[\mathbf{E}_{q_{\phi }(z|x)}[log p_{\theta }(x|z)] ]
\end{aligned}
\end{equation}

$\beta$-VAE is a modification of the VAE framework that introduces an adjustable hyperparameter $\beta$ to the original VAE objective: 
\begin{equation}
\mathcal{L} = \mathbb{E}_{q_{\phi }}(log p_{\theta}(x|z))-\beta D_{KL}(q_{\phi }(z|x)|| p_{\theta}(z))
\end{equation}
Well chosen values of $\beta$ (usually $\beta>1$) result in more disentangled latent representations z. 
When $\beta$ = 1, the $\beta$-VAE becomes equivalent to the original VAE framework. It was suggested that the stronger pressure for the posterior $q_{\phi}(z|x)$, to match the factorized unit Gaussian prior p(z) introduced by the $\beta$-VAE objective, puts extra constraints on the implicit capacity of the latent bottleneck z. Higher values of $\beta$ necessary to encourage disentangling often lead to a trade-off between the fidelity of $\beta$-VAE reconstructions and the disentangled nature of its latent code z (see Fig. 6 in [15]). This is due to the loss of information as it passes through the restricted capacity latent bottleneck z.

\subsection{Adversarial attacks}
Evasion attacks have long been studied on machine learning classifiers \cite{barreno2010security,lowd2005adversarial}, and are practical against many types of models \cite{biggio2013evasion}. These evasion attacks over neural networks are referred to as adversarial examples \cite{szegedy2013intriguing}. Namely, for a given input sample $x$, the adversarial example is a sample $x'$ that is similar to x (according to particular measure metrics) but so that the classifier's decision $C(x) \neq C(x')$ \cite{biggio2013evasion}. 
A classifier can misclassify an adversarial example for two reasons. (1) The adversarial example is far from the boundary of the manifold of the task. For example, the task is a handwritten digit classification, and the adversarial example is an image containing no digit, but the classier has no option to reject this example and is forced to output a class label. (2) The adversarial example is close to the boundary of the manifold. If the classier poorly generalizes the manifold in the vicinity of the adversarial example, then misclassification occurs.

Let $ \mathbb{U}$ be the set of all instances in the sample space. A classification function is denoted by $C$, which outputs for each instance $x \in \mathbb{U}$ a predicted class $C(x)=y$. Let $ \mathbb{Y}=\{y_1, \cdots , y_m \}$ denote the set of classes for a certain classification task. 
Each classification function assumes a data generation process that produces each instance $x \in \mathbb{U}$ with probability $p(x)$. Let $ \mathbb{N}$ be a manifold that consists of instances that act naturally with regard to a certain classification task, following a data generation process. $\mathbb{N}$ can be approximated by a set of natural instances for a classification task \cite{meng2017magnet}, e.g., MNIST.
The goal of the adversarial example is to find certain perturbation on $x$ to generate adversarial example $x^* \in \mathbb{U}\setminus \mathbb{N}$ that fools a specific $C$ to misclassify, i.e. $ C(x^*) \neq C(x) $. 

The adversary is assumed to have the knowledge of the original classifier but is not aware of the detector and purifier. Therefore, the goal of the adversary is only to fool the unsecured classifier. 
\subsection{Adversarial defenses}
Defense on neural networks is much harder compared with attacks. We summarize some ideas of current approaches to defense and compare them to our work. 

\subsubsection{Adversarial Training}
One idea of defending against adversarial examples is to train a better classifier \cite{shaham2015understanding}. An intuitive way to build a robust classifier is to include adversarial information in the training process, which we refer to as adversarial training. For example, one may use a mixture of normal and adversarial examples in the training set for data augmentation \cite{szegedy2013intriguing}, or mix the adversarial objective with the classification objective as regularizer \cite{goodfellow6572explaining}. Although this idea is promising, it is hard to reason about what attacks to train on and how important the adversarial component should be. Currently, these questions are still unanswered. 

\subsubsection{Defensive Distillation} 
Defensive distillation \cite{papernot2016distillation} trains the classifier in a certain way such that it is nearly impossible for gradient-based attacks to generate adversarial examples directly on the network. Defensive distillation leverages distillation training techniques \cite{hinton2015distilling} and hides the gradient between the pre-softmax layer (logits) and softmax outputs. However, \cite{carlini2017towards} showed that it is easy to bypass the defense by adopting one of the three following strategies: (1) choose a more proper loss function (2) calculate gradient directly from pre-softmax layer instead of from post-softmax layer (3) attack an easy-to-attack network first and then transfer to the distilled network. 
\subsubsection{Detecting Adversarial Examples} 
The detection-based defense against adversarial examples for a classifier $C$ aims to establish a detector $d_C: \mathbb{U} \rightarrow \mathbb{Y} \cup \{J\}$. $J$ is the judgment that the input is unlikely from the manifold of the normal instances. 
Further, the purification-based defense is to build a purifier $p: \mathbb{U} \setminus \mathbb{N} \rightarrow \mathbb{N} $ to reconstruct suspicious instances with small distortion only using some essential features, in order to move adversarial examples towards the manifold of normal examples.
The defense aims to increase the accuracy of the classifier with the presence of adversarial examples by (1) detecting the input as an adversarial example or a normal image while rejecting suspicious instances with huge distortion; (2) purifying suspicious instances with small distortion by reconstruction.
One strong defense is to detect adversarial examples with hand-crafted statistical features \cite{grosse2017statistical} or separate classification networks \cite{metzen2017detecting}. A representative work of this idea is \cite{metzen2017detecting}. For each attack generating method considered, it constructed a DNN classifier (detector) to tell whether an input is normal or adversarial. The detector was directly trained on both normal and adversarial examples. The detector showed good performance when the training and testing attack examples were generated from the same process, and the perturbation was large enough, but it did not generalize well across different attack parameters and attack generation processes.

\section{Defense via Semantic Feature Manipulation}

\subsection{ FM-Defense overview}
We propose Feature Manipulation based defense (FM-Defense), a framework for defending adversarial examples via detecting and purifying. 
\begin{figure}[!htb]
\setlength{\abovecaptionskip}{-0.05cm}
\setlength{\belowcaptionskip}{-0.2cm}
\includegraphics[width=3in,height=3in]{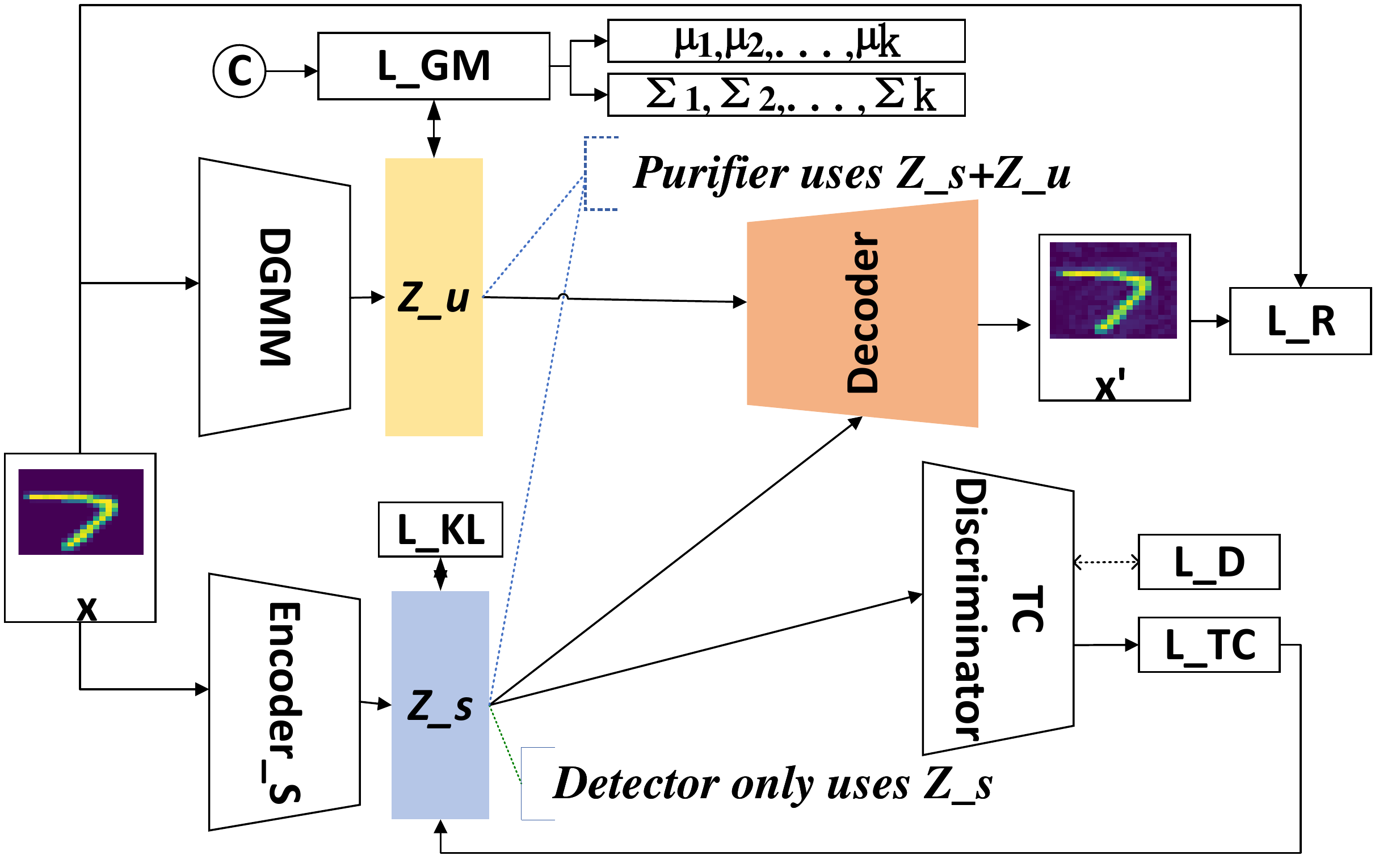}
\centering
\caption{Scheme of FM-Defense.}
\end{figure}

Three key challenges need to be addressed by FM-Defense:
(1) How to make the feature manipulation feasible in an interpretable manner?
(2) How to improve the discrimination ability of the detector to recognize adversarial examples by manipulating features efficiently?
(3) How to enhance the quality and completeness of the purified instances by reconstruction? 
To address these three issues, FM-Defense is composed of three components: 

 (1) \textit{Representation with improved disentanglement.} We initially train a disentangle representation model, combo-VAE, enhanced from $\beta$-VAE \cite{higgins2017beta}, on a clean dataset. The objective here is to make disentangled latent codes that have semantic meaning so that they are easy to be controlled. Such models consist of two components: 
the encoder $E : X \rightarrow Z$ takes the input $x \in R^d$ and maps it to $z$ (the latent variable of the bottleneck vector). The decoder $D : Z \rightarrow X’$ tries to reconstruct the input data from $z$. The encoder can map high-dimensional input instance $x$ to disentangled low-dimensional latent codes $z$, i.e., the one latent code can only control one certain feature. The decoder is used to reconstruct the input from the low-dimensional latent code $z$. 
For simple images, e.g. $28\times28$ handwritten digits, the latent code can be a $m$-dimensional vector. For complex images, e.g. $128\times 128$ face images, the latent codes are $m$ channels of $n \times n$ feature maps. 
We find that some $n \times n$-dimensions feature map can reveal some disentangled semantic features as well. 
Therefore, each channel of $n \times n$ feature map can be considered as one latent factor used for manipulation for simplicity, namely, each element of the $n \times n$ feature map is simultaneously changed at the same scale. 
 The combo-VAE to extract the disentangled latent codes for each instance so that it is feasible to select and manipulate a number of codes that reveal the desired semantic features, e.g. thickness of the digit. The feasibility of feature manipulation is related to the disentanglement level. Therefore, strategies are used to improve the disentanglement, as described in the following sections.

  (2) \textit{Detector with fine discrimination ability}. Given an instance, we first vary a latent code $i$ for $T$ times to obtain $n$ morphs reconstructed by the decoder. We then record the ratio of unchanged classification prediction by applying a certain classifier (to be protected) on these $n$ morphs compared with the original prediction. The ratio is used as a resistance indicator $r^{(i)}$ for code $i$. At one time, we change one of $m$ selected latent codes in turn and obtain a $m$-dimensional resistance vector $\overrightarrow{r}$ for each instance. We find the resistance ability of normal instances is significantly better than that of adversarial ones, as shown in the histogram of Figure 1. Therefore, a $m$-dimensional threshold configuration $\overrightarrow{\theta_r}$ for all selected $m$ latent codes can be decided on the normal instance to distinguish normal and adversarial instances. An instance, that meets $r^{(i)} > \theta_r^{(i)},\  \forall\ r^{(i)} \in \overrightarrow{r}$, will be recognized as normal. Otherwise, it is recognized as suspicious.
 
 (3) \textit{Purifier with comprehensive reconstruction ability}. 
It is feasible to decide a threshold that can achieve nearly $100\%$ adversarial detection accuracy (True Positive). However, this will cause a large number of normal instances to be improperly recognized as adversarial, i.e., high False Positive ratio. 
We assume there exists a manifold of resistance on clean instances. Therefore, we use another threshold $ \theta_d$ over the distance $Dis(.)$ between the resistance vector of an instance $\overrightarrow{r}$ and the resistance threshold vector $ \overrightarrow{\theta_r}$, to build a salvageable set consists of suspicious instances close to the manifold of normal instances.
\begin{equation}
Dis(\overrightarrow{r},\overrightarrow{\theta_r}) = \sum |r^{(i)}-\theta_r^{(i)}|, \ \forall  r^{(i)} < \theta_r^{(i)}
\end{equation}
\begin{figure}[!htb]
\centering
\setlength{\abovecaptionskip}{-0.05cm}
\setlength{\belowcaptionskip}{-0.2cm}
\includegraphics[width=3in,height=2in]{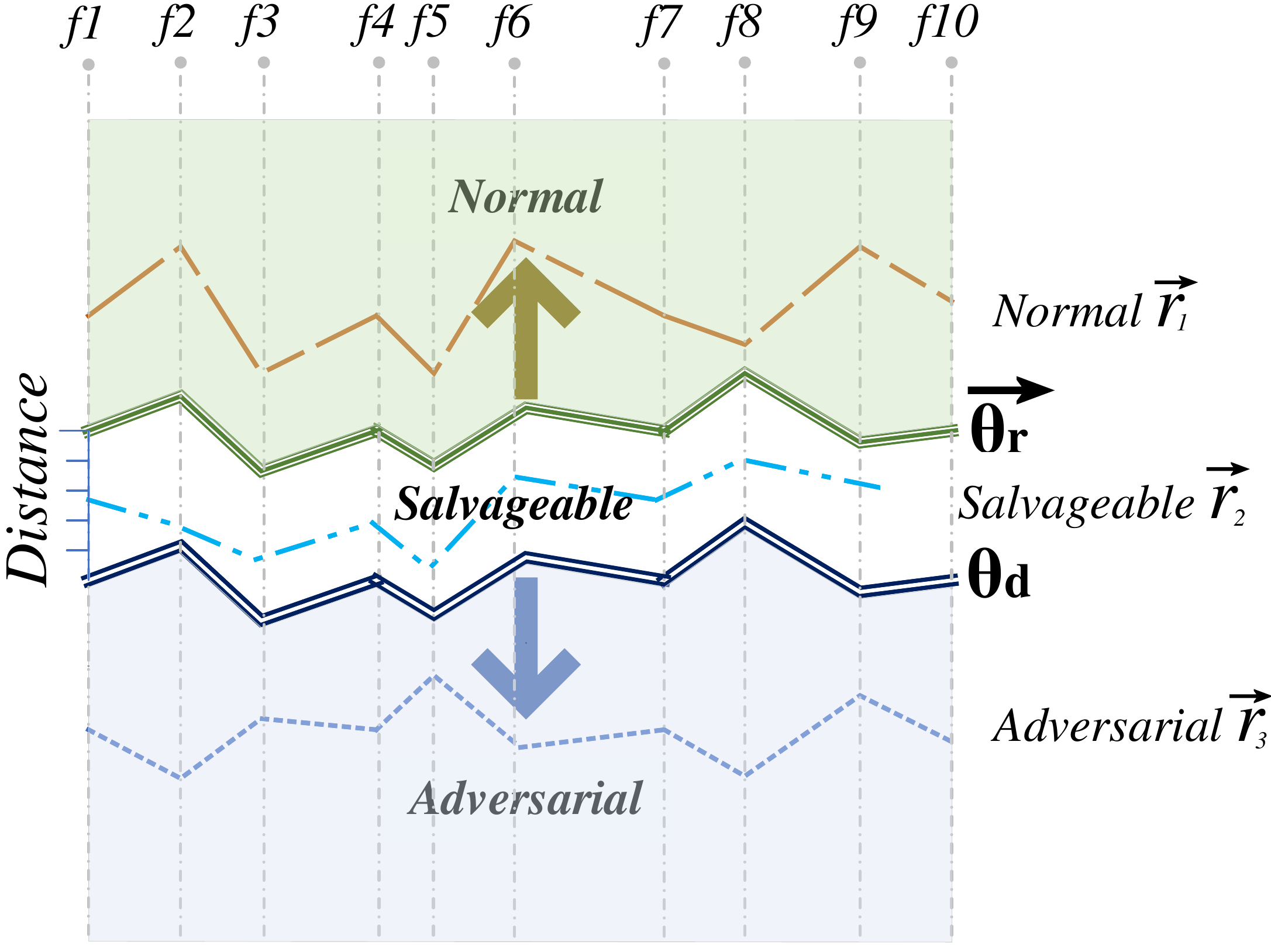}
\centering
\caption{Threshold settings.}
\end{figure}

 As illustrated in Figure 3, if a suspicious instance from the detection has a distance of more than $ \theta_d$, then it will be rejected as adversarial. Otherwise, it is incorporated in a salvageable set that will be reconstructed by a combo-VAE based purifier. The purifier moves suspicious examples in the set towards the manifold of normal examples to correctly classify adversarial examples with small distortion or clean instances improperly recognized as adversarial. Namely, the purifier reconstructs close approximation on the normal manifold before feeding to the target classifiers. 
To improve the quality of reconstructed instance, we enhance the combo-VAE to absorb both significant class-unique features and class-shared features. 
Details of these components are given in the following sections.
\subsection{Representation with improved disentanglement}
VAE-based autoencoders and their variations are commonly applied for disentanglement learning. Specifically, the encoder $E$, parameterized by $q_{\phi }(z|x)$, is trained to convert high-dimensional data $x$ into the latent representation bottleneck vector $z$ in the latent space that follows a specific Gaussian distribution $p(z) \sim N(0, 1)$. The decoder $p_{\theta}(x|z)$ is trained to reconstruct the latent vector $z$ to $x$. The encoder and decoder are trained simultaneously based on the negative reconstruction error and the regularization term, i.e., Kullback-Leibler (KL) divergence between $q_{\phi }(z|x)$ and $p(z)$. The regularization term is used to regularize the distribution $q_{\phi } (z|x)$ to be Gaussian distribution whose mean $\mu$ and diagonal covariance $\sum$ are the output of the encoder. 
We apply a combo-VAE to get good disentanglement in z by improving the inner-independence of latent codes. 
Specifically, Total Correlation (TC) \cite{watanabe1960information} is used to encourage independence in the latent vector $z$, as given in Equation 2. 
\begin{equation}
TC(z) = KL(q(z)||\bar q(z)) = Eq(z)[log\frac{q(z)}{\bar q(z)}] 
\end{equation}
As TC is hard to obtain, the approximate tricks used in \cite{kim2018disentangling} is applied to estimate TC. Specifically, a discriminator $D_tc$ is applied to classify between samples from $q(z)$ and $\bar q(z)$. Thus learning to approximate the density ratio is needed for estimating TC \cite{kim2018disentangling}. $ D_{tc} $, parameterized by $\upsilon $, is trained with other components jointly. Thus, the TC term is replaced by the discriminator-based approximation as follows:
\begin{equation}
TC(z) \approx E_{q(z)}[log \frac{D(z)}{1-D(z)}]
\end{equation}
The objective of combo-VAE is augmented with a TC \cite{watanabe1960information} term to encourage independence in the latent factor distribution as follows:
\begin{equation}
\begin{aligned}
E_{ q_{\phi } (z|x^{(i)})}[log p_{\theta} (x^{(i)}|z) - L_{KL}( q_{\phi } (z|x^{(i)})||p(z))]- \gamma L_{TC}
\end{aligned}
\end{equation}
Note that this is also a lower bound on the marginal log-likelihood $E_{p(x)}[log p(x)]$. 
The first part reveals the reconstruction error, denoted by $L_R$, evaluating whether the latent bottleneck vector z is informative enough to recover the original instance. $L_R$ can be defined as the $l_2$ loss between the original instance and the reconstructed instance. 
The second part is a regularization term, denoted by $L_{KL}$, to push $ q_{\phi } (z|x)$ to match the prior distribution $p(z)$.
The third part is the TC term, denoted by $L_{TC}$, to measure the dependence for multiple random variables. 

As shown in Figure 2, the parameter $\phi $ of encoder $ q_{\phi } (z|x)$ is then trained by $L_{KL}$, $L_R$ and $L_{TC}$ in terms of $-\nabla_{ \phi }( L_{KL} +L_{R}+ \gamma L_{TC}) $. 
The parameter $\theta $ of decoder is updated in terms of $-\nabla_{\theta }(L_R) $. 
The parameter $ \upsilon $ of TC-discriminator is updated in terms of $-\nabla_{\upsilon }(L_T) $, i.e. $-\nabla_{\upsilon }\frac{1}{2|B|}[\sum_{i \in B}log (D_\upsilon (z^{(i)})+\sum_{i \in B'}log (1-D_\upsilon (permutedim(z'^{(i)}))] $. Here, the permutedim function is to random permutate on a sample in the batch for each dimension of its $z$, similar to \cite{kim2018disentangling}.

\subsection{Detector with fine discrimination}
The indicator for adversarial detection should easily differentiate normal and adversarial instances, be feasible and stable to conduct, and attack-agnostic. The classification accuracy resistance over the morphs, derived from feature manipulation by changing a certain latent code, can meet these criteria. The discrimination ability of the detector depends on the ability to reduce the false-positive ratio (normal instances to be recognized as adversarial) and the naturality of the morphs. Consequently, we apply two strategies: normal value range selection and natural morph generation. 

\textit{Natural morph generation.} 
 The initial step is to find the normal value range of each code on the clean validation set, then the morphs are produced via manipulating each code within its normal value range. As the latent codes are disentangled, independent (all from $N(0,1)$) and have semantic meaning, some latent codes that reveal non-significant intrinsic features (e.g., thickness for handwritten digits) will be selected and their normal range can be decided empirically in an interpretable manner for human on a validation set. 
 To obtain the morphs by feature manipulation, we can incrementally add/reduce a fixed value on the original learned latent codes within the normal range. However, the modified latent vector maybe not be on the manifold of normal instances. If that happens, an unnatural instance will be reconstructed by the decoder. Hence, we conduct an iterative stochastic search to make the morphs on the manifold by adding natural noise. Specifically, we increase the search range by $\Delta r$ within which the perturbation for a certain latent code $\Delta z_i$ is randomly sampled ($B$ samples for each iteration) until we produce $N$ natural latent code with the value in the normal value range to reconstruct $N$ natural morphs. We then evaluate the resistance on classification for this latent code using the targeted classifier. Iterative, we can get a $m$-dimensional resistance evaluation vector for each instance.
 
 \textit{Threshold selection strategies.} 
 Given a targeted classifier, we decide a resistance threshold for each latent code on a validation set containing only clean instances. 
 The threshold of resistance is decided for each factor so that the false-positive rate on the validation set is below a pre-defined value $1-\rho$ (i.e. more than $\rho\%$ clean instance are correctly recognized). Therefore, we can decide a unified and fixed resistance threshold for all latent codes or formulate a specific threshold for each latent code in terms of $\rho$.

\subsection{Purifier with comprehensive reconstruction}
We assume that suspicious instances with a small distance between their resistance vector and the resistance threshold vector can be considered as close to the manifold of the normal instances. Namely, the distance is under the $\theta_d$, which is set as the $\eta\%$ fractile on clean validation data. Therefore, a VAE-based purifier is used to reconstruct and move them towards the manifold of normal examples. 
In regular VAEs, the prior over the latent variables is commonly an isotropic Gaussian, resulting in limited representation because the learned representation can only be unimodal and does not allow for more complex representations \cite{dilokthanakul2016deep}. Consequently, the regular VAE-based disentangled learning generally learn some class-shared information and with some essential class-unique information lost. 
This limitation causes only the suspicious instances with tiny perturbation, while suspicious instances with larger perturbation, e.g. derived from FGSM attack in Figure 6, can not be reconstructed correctly. 
To address the completeness and accuracy of the VAE-based purifier, we enhance the combo-VAE by incorporating class relevant conditional information to guide the reconstruction. Our latent codes consist of two components: class-unique representation, $z_u$ (e.g., important features unique to each digit), and commonly shared across all classes, $z_s$ (e.g., the thickness of handwriting digits). As shown in Figure 2, the enhanced combo-VAE has a similar scheme to a VAE, but instead of using exclusively the same data for the input and output of the network, we use class-unique additional information as an extra input to the decoder. 

Specifically, we assume the observed instances are derived from a mixture of Gaussians, i.e. the inference of the class of an instance is equivalent to inferring which mode of the latent codes $z_u$ of the data point was generated from. Namely, we use a mixture of Gaussians as our prior for the $z_u$, used as conditional information for training decoder of VAE. 
For each class label $c$, we assume it has $K$ features such as different writing styles for handwritten digits, namely $K$-dimensional $z_u$. 
Therefore, we first train a deep neural network using Gaussian Mixture loss (DGMM) that maps input $x$ to $z_u$ that is learned with the supervision of the categorical class label $c$. $z_u$ reveals $K$ features for the label of a given input instance. The DGMM is solely trained using the clean labeled instances, and the output is a logistic regression on the latent representation of $K$ features, as a classification task.  

Each feature $ z_u^{(k)}$ follows a mixture of $K$ Gaussian distribution with learned mean $\mu_c$ and covariance $\Sigma_c$ for each class $c$, given by neural networks of DGMM with parameters $ \kappa_\mu $ and $ \kappa_\Sigma $ respectively. 
\begin{equation}
p(z_u^{(k)})=\sum_{c}^{C}\emph{N}(c;\mu_c,\Sigma_c)p(c)  
\end{equation} 
Here, $p(c)$ is the prior probability of class $c$. 
The loss of DGMM is calculated as the cross-entropy between the posterior probability $q(c|z_u)$ and the corresponding one-hot class labels, denoted $L_{cls}$, combined with a likelihood regularization term to force the training samples to obey the assumed GM distribution, denoted $L_{lkd}$ \cite{dilokthanakul2016deep,zheng2019disentangling}. 
$L_{cls}$ can let $z_u$ contain as much label information as possible, as the MI between $z_u$ and class $c$ are added to the maximization objective function.
\begin{equation}
\begin{split}
L_{cls} = -Eq_{\kappa} (z_u|x)\sum_c I(c = y) log\  q(c|z_u)\\= -log\frac{N(z_u;\mu_y,\Sigma_y)p(y)}{\sum_k N(z_u;\mu_k,\Sigma_k)p(k)}
\end{split}
\end{equation}
$L_{lkd}$ is applied for measuring to what extent the training samples fit the assumed distribution, which can be simplified as Equation 7 when $p(c)$ is simply set to $1/C$ for all classes. The $L_{lkd}$ for a given class $c$ is given as follows:
\begin{equation}
L_{lkd}= -logN(z_u;\mu_c,\Sigma_c)
\end{equation}
Consequently, the loss for $DGMM$ is 
\begin{equation}
L_{GM} = L_{cls} +\lambda_{lkd}L_{lkd},
\end{equation}
where $\lambda$ is a non-negative weighting coefficient. 

In addition, an encoder $E_s$ is trained to map input $x$ to $z_s$ where each code is forced to follow the standard Gaussian $N(0, I)$, implemented completely by the $ q_{\phi } (z|x)$ in Section \textit{Representation with improved disentanglement}. 
The input of a pre-trained DGMM can a given instance $x$ without a label, since DGMM will output the $K$-dimensional feature vector of the most likely class of $x$. 
The latent codes $z_s$ and $K$-dimensional $z_u$ derived from the pre-trained DGMM are then simply concatenated together to a decoder to reconstruct the input $x$. 
The loss of decoder is used to measure how probable it is to generate $x$ by using the distribution $p(x’ | z_u, z_s)$, that is, is a distance between $x$ and reconstructed $x’$.
  
The training of combo-VAE is two-stages. 
Initially, the DGMM (modeled by $q_{\kappa}$) is updated using $L_{GM}$ to learn mean $\mu_c$ and covariance $\Sigma_c$ of the prior $p(z_u|c)$, encouraging $z_u$ to be label dependent and follows a learned Gaussian mixture distribution. At the second step, the encoder $E_s$ and the decoder are trained jointly to reconstruct images with concatenated $z_s $ and $z_u$. The Encoder $E_s$ (modeled by $q_{\phi}$) is intended to extract class-shared code $z_s$. This is trained by $L_{kl}$ and $L_{R}$ to make $z_s$ be close to $N(0,1)$. 
Then the decoder $p_{\theta}$ generates a reconstruction image using the combined feature of $z_s$ and $z_u$ with the loss $L_{R}$.

\section{Experiments}
The performance of FM-Defense is evaluated against the state-of-the-art adversarial attacks on two datasets: MNIST \cite{lecun1998gradient}, FMNIST (2D shapes) \cite{xiao2017fashion} and CelebA \cite{liu2015deep} (Face). 
The adversaries are assumed have no knowledge about the detector and purifier (black-box setting). They only focus on generating adversarial examples that aim to maximize the prediction errors on a target classifier, and do not care which class the victim classifier outputs as long as it is different from the ground truth. 

\subsection{Setup}

On MNIST, FMNIST and CelebA, 20,000 clean examples are chosen for training one-off combo-VAE. We randomly select 5000 clean images (named CLE, labeled 0) and the corresponding 5000 adversarial examples (named ADV, labeled 1), respectively. These datasets are used to test the efficiency of the FM-Defense. Besides, another 2000 clean instances are chosen as the validation data (named VAL) to decide thresholds. 
We trained a classifier for MNIST using the setting in \cite{carlini2017towards} with an accuracy of  $99.4\%$.  We train a classifier using the setting in \cite{he2016deep} for FMNIST. For the CelebA, we train an identification classifier using the setting in \cite{simonyan2014very} with an accuracy of $ 94.7 \%$.
We normalized the data between 0 and 1 instead of [0, 255] for simplicity. Table I shows the architectures of the combo-VAE for MNIST, FMNIST, and CelebA.

\begin{table*}[!htb]
\centering
\setlength{\abovecaptionskip}{-0.05cm}
\setlength{\belowcaptionskip}{-0.2cm}
\caption{The network structures. N=Neurons, K=Kernel size, S=Stride size. Convolutional layer is denoted by C. or DC. The residual basic block is denoted as RSB.}
\centering
\label{tab:my-table}
\begin{tabular}{|c|l|c|l|}
\hline
\textbf{\textbf{Encoder/DGMM}} & \textbf{Face Encoder} & \textbf{\textbf{Decoder}} & \textbf{Face Encoder} \\ \hline
C.ReLU N32,K4,S2               & C.LReLU N64,K7,S1     & Dense.ReLU 128/512        & RSB. N512,K3,S1       \\
C.ReLU N32,K4,S2               & C.LReLU N128,K3,S2    & Dense.ReLU N64,K4         & RSB. N512,K3,S1       \\
C.ReLU N64,K4,S2               & C.LReLU N256,K3,S2    & C.LReLU N64,K4,S2         & RSB. N512,K3,S1       \\
C.ReLU N64,K4,S2               & RSB. N512,K3,S1       & C.LReLU N32,K4,S2         & DC.LReLU N256,K3,S2   \\
Dense 128                      & RSB. N512,K3,S1       & C.LReLU N32,K4,S2         & DC.LReLU N128,K3,S2   \\
Dense 2*10                     & RSB. N512,K3,S1       & C.LReLU N1,K4,S2          & DC.LReLU N3,K7,S1     \\ \hline
\end{tabular}
\end{table*}

We use fast gradient sign method (e.g., FGSM ($\epsilon=0.3,  L^{\infty}$) \cite{kurakin2016adversarial}), DeepFool ($L^{\infty}$) \cite{moosavi2016deepfool,carlini2017towards}, and CW ($L^{2}$) attack \cite{moosavi2016deepfool,carlini2017towards} for the experiments, implemented using Fooling box \cite{rauber2017foolbox} .
We select 20-dimensional latent codes to manipulate MNIST and 10-dimensional for FMNIST. The face image will be mapped into 256 channels of 64*64 feature maps, and each channel is considered as one latent factor. 
We generate $ 100$ morphs for each latent factor of instance by varying each latent code for 100 times within according normal value range. The morphs were fed into the target classifier to calculate the resistance vector for each instance. 
For MNIST and FMNIST, we decide the resistance threshold vector, respectively, such that the false-positive rate of the detector on the validation set VAL is at most $0.001$ for all sleeted latent codes. This means each detector mistakenly rejects no more than $0.1 \%$ clean instances in the validation set, i.e., $\rho=99.9\%$. Besides, distance threshold values are decided for MNIST and FMNIST, respectively, set as the $\eta=90\%$ fractile on clean validation data in our experiments. 
Other default hyper-parameters are given as follows: $\gamma=40,\ \lambda_{lkd}=0.1$. 
For the face images of CelebA, we find that only one randomly selected 64*64 feature map as a latent factor can achieve well-detecting performance, approximately $100\%$, even without the TC-optimization mentioned in Section 2.3. Therefore, a basic combo-VAE without optimization strategies is used for face images and only one latent factor (64*64 feature map) is used for detection.
\subsection{Defense evaluation against adversarial attacks}
\subsubsection{Detection accuracy evaluation.}
We first evaluate the performance of the detector for varying resistance thresholds. The overall results on MNIST are shown in the first row of Figure 4 in terms of adversarial, clean and overall detection accuracy. Here, we test on both CLE and ADV datasets, respectively. The correct decision is that adversarial examples are recognized as 1 while normal ones as 0. Initially, we use a unified resistance threshold for all selected latent codes. 
The adversarial detection accuracy (legend as Adv) is the proportion of adversarial instances in ADV to be recognized as adversarial, i.e. True-Positive. The clean detection accuracy (legend as Clean) is the proportion of clean instances in CLE to be recognized as clean, i.e. True-Negative. The overall detection accuracy (legend as Overall) is the proportion of all correctly detected instances in both ADV and CLE. We observe that even for a small resistance threshold, e.g. $5$, it can detect more than $75\%$ adversarial examples, and retaining $81\%$ clean instances correctly labeled. The adversarial detection accuracy of FM-Defense on ADV is above $99.9\%$ for both MNIST and FMNIST on all the potential attacks, including CW attack ($99.9\%$ adversarial detection accuracy at united resistance threshold $85$ with $13\%$ correctly recognized clean instances). Note that we achieved such high accuracy without any adversarial examples required and only based on threshold vectors that are easy to be decided experimentally. 

\begin{figure*}[!htb]
\setlength{\abovecaptionskip}{-0.05cm}
\setlength{\belowcaptionskip}{-0.2cm}
\includegraphics[width=5.5in,height=4.2in]{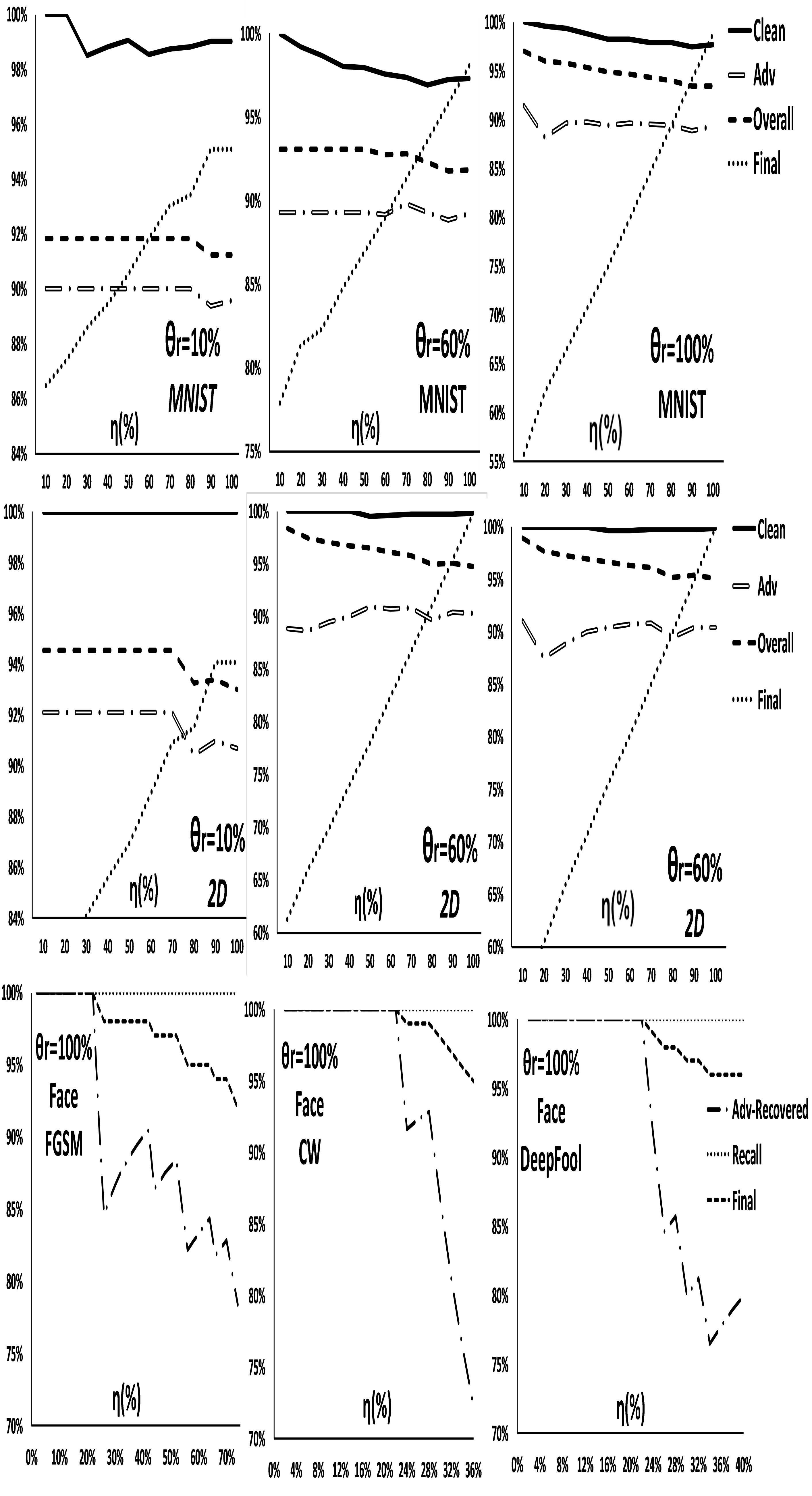}
\centering
\caption{Detection accuracy evaluation on MNIST, FMNIST and CelebA, 2D/Face is correspond to FMINST/CelebA respectively.}
\end{figure*}

However, the False-Positive (FP) rate of normal instances increases as the resistance threshold and adversarial detection accuracy rise. The major concern here is how to balance the adversarial detection accuracy and the FP rate. Combining the correctly recognized normal instances and adversarial instances together, the overall detection accuracy reaches the best at nearly $80\%$ on the threshold at $30$, which can be used as a balanced threshold, with $85-93\%$ adversarial detection accuracy and $75\%$ clean detection accuracy. These results demonstrate that our method can efficiently thwart adversarial examples while achieving an acceptable FP rate.

We also demonstrate the impacts of choosing different threshold selection strategies, i.e., varying the $\rho$ to decide a specific resistance threshold for each latent code and selecting different codes. As shown in the second row of Figure 4, both adversarial detection accuracy and FP rate (1-clean detection accuracy) increase as the $\rho$ rises. Hence, it is feasible to decide a $\rho$ that can balance the adversarial detection accuracy and FP rate. Overall, the specific resistance threshold strategy performs better than the unified threshold, with relatively high adversarial detection accuracy. 
In addition, we find that selecting different latent codes affects detection accuracy. As shown in the third row of Figure 4, selecting latent codes for FM-Defense against the CW attack over MNIST has a various discriminatory ability. It is feasible to decide an efficient code selection configuration experimentally. 
The third row of Figure 4 are results on FMNIST(2D) dataset, which confirms the findings mentioned above. 
The last row is the resistance statistics on face images. For complex image, it is interesting to find that the clean images are more resistant to the latent factor change within a certain value range, reaching to $100\%$. In contrast, the adversarial ones are more sensitive to such change, totally below $80\%$. Therefore, it is feasible to use a higher resistance, e.g. $90\%$, to detect all adversarial examples (detection accuracy = 100\% and FP rate=0).  

\subsubsection{Purification accuracy evaluation}
We then evaluate the purification performance of our approach on the detected suspicious instances. First, we calculate the distance between the resistance vector of a given instance and the resistance threshold vector to establish a salvageable set for purification. The set includes only adversarial examples and misclassified clean instances. 

We measure the recall in Figure 5, i.e., the proportion of detected suspicious instances in the salvageable set that is regained the correct labels after reconstruction, as vary the distance threshold $\theta_d$ (set at the fractile of $\eta \%$, i.e. how many salvageable instances are purified) as well as resistant threshold $\theta_r$ (the unified threshold for simplicity). 
There are three types of recall: clean recall, adversarial recall, and overall recall to reveal the correctly recovered proportion on ADV, CLE, and combination. We also record the final accuracy of FM-Defense, namely $True Positive+True Negative+\frac{Recovered}{Salvageable}$. 
The first row of Figure 5 is the results on MNIST. When $\theta_r$ is $60\%$ or $100\%$, more than $95-98\%$ adversarial instances are recognized while about $80\%$ clean instances are wrongly labeled.  As $\eta$ (i.e. $\theta_d$) rises, the adversarial and clean recall after purification slightly decreases but retains a high level (more than 95$\%$). Nearly $90\%$ adversarial instances and  $100\%$ clean instances in the salvageable set are recovered to the correct label when the $\eta$ is $10\%$.
The final accuracy of FM-Defense increases as the $\eta$ goes up, reaching $99\%$ when the distance threshold is $100$ (all salvageable instances are purified). In addition, all the recall values and final accuracy increase when rising $\theta_r$. This reveals that we can set a high resistance threshold to make all adversarial instance to be detected without concerns about the FP ratio since our purifier can well recover both the adversarial instances and wrongly recognized clean ones in the suspicious set. 
The experiments on FMNIST (2D) demonstrate the same conclusions, as shown in the second row of Figure 5. 
\begin{figure*}[!htb]
\centering
\setlength{\abovecaptionskip}{-0.01cm}
\includegraphics[width=5.5in,height=5in]{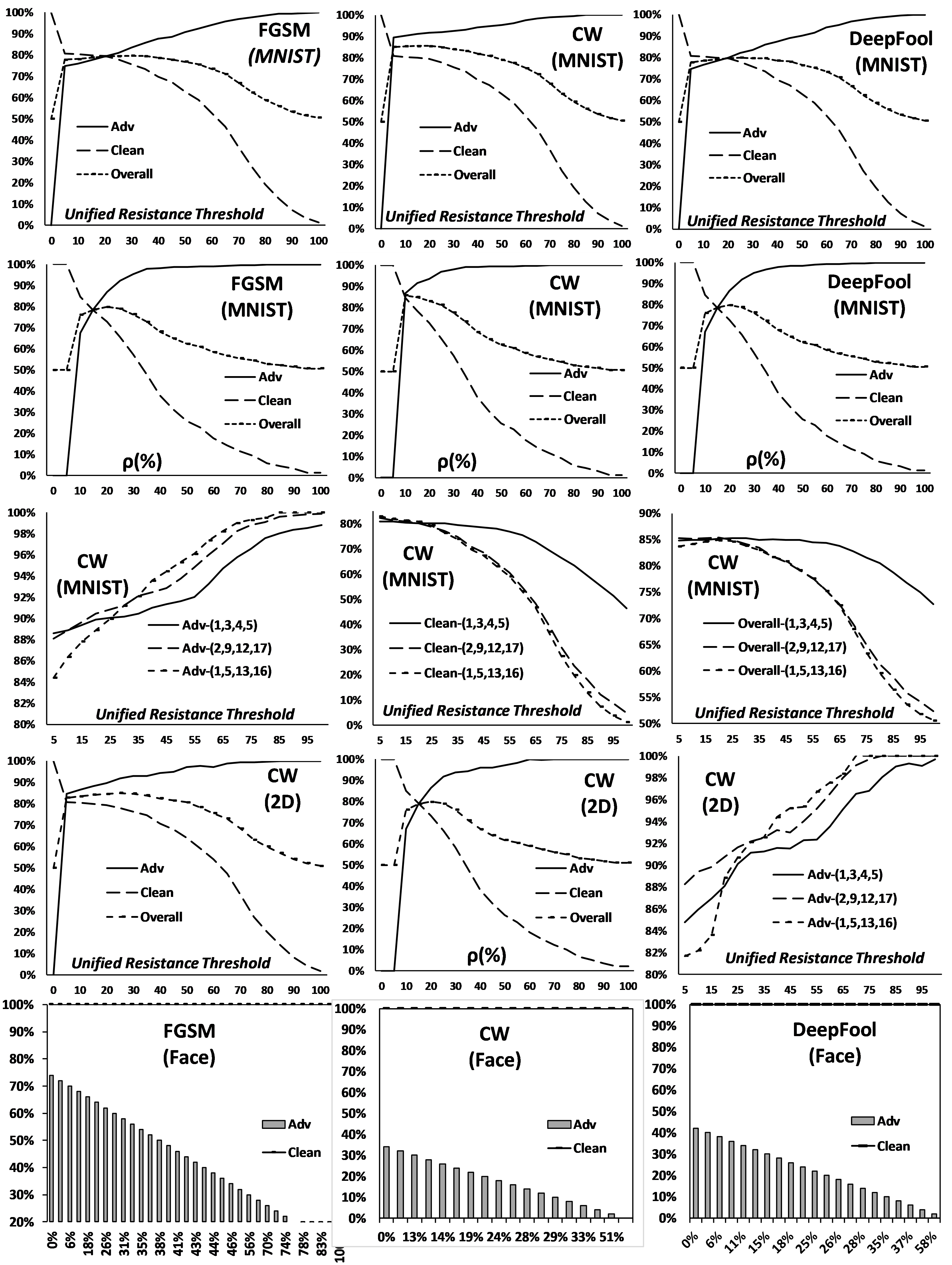}
\centering
\caption{Purification results for CW attack with varying distance threshold.}
\end{figure*}

The last row illustrates the performance on the face image. As the FP rate is 0, namely all clean images are recognized as normal and total recall of detection is $100\%$. Consequently, the key point here is to select the proportion of the suspicious (adversarial) images for purification, i.e. $\eta$. As shown, it feasible to choose a suitable $\eta$ to achieve a $100\%$ final accuracy. However, the final accuracy and the percentage of selected adversarial images recovered to a normal decrease when increasing the $\eta$. The reason is that some adversarial examples have impacted the significant characters of identification. 

These results confirm that our detector and purifier can efficiently recognize adversarial examples that far from the manifold of normal instances. They also can effectively move the misclassified clean instances and the salvageable adversarial examples towards the normal manifold.
\subsubsection{Purification quality evaluation}
We further illustrate the quality of the purified instances through reconstruction by our combo-VAE, and compare it with other VAE-based generative models, e.g. Factor-VAE \cite{kim2018disentangling}. We show the purified images on the MNIST and CelebA datasets in Figure 6. It reveals that samples reconstructed by ours can retain more class-unique features on clean instances compared with Factor-VAE. The performance of ours on reconstructing adversarial examples is much better than Factor-VAE, as DGMM is used to capture more class relevant information that can guide reconstruction. Therefore, our defense has better generalization ability and can purify more plentiful adversarial examples instead of only these small distortion. 
\begin{figure*}[!htb]
\centering
\setlength{\abovecaptionskip}{-0.05cm}
\setlength{\belowcaptionskip}{-0.2cm}
\includegraphics[width=5.5in,height=2.2in]{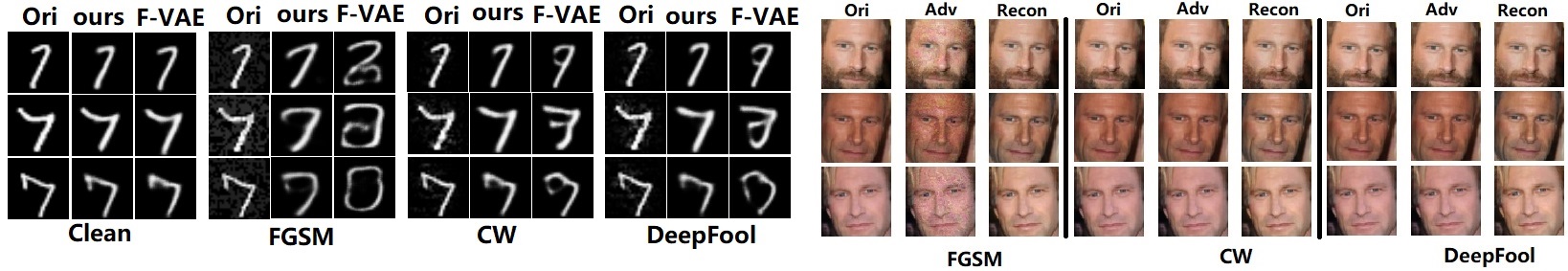}
\centering
\caption{Purification quality illustration.}
\end{figure*}

\subsubsection{Overall evaluation with comparisons}
Table II shows the effects of our defense against different adversarial attacks on the MNIST and CelebA, comparing with the state-of-the-art defenses, e.g., MagNet, Defense-GAN and FBGAN. 
\begin{table*}[!htb]
\centering
\setlength{\abovecaptionskip}{-0.03cm}
\setlength{\belowcaptionskip}{-0.1cm}
\caption{Classification accuracy ($\%$, MNIST/CelebA) under different attack and defense methods}
\label{tab:my-table}
\begin{tabular}{|l|l|l|l|l|l|}
\hline
\textbf{\begin{tabular}[c]{@{}l@{}}Accuracy\end{tabular}} & \textbf{No defense} & FM-Defense & Defense-GAN & MagNet    & FBGAN       \\ \hline
FGSM                                                                         & 21.2/7.6            & 99.8/99.9  & 83.21/32.17 & 74.6/59.2 & 80.43/33.71 \\ \hline
CW                                                                           & 0/0                 & 98.78/98.9 & 80.11/28.22 & 19.6/40.5 & 90.8/35.5   \\ \hline
DeepFool                                                                     & 9.1/2.1             & 99.7/99.8  & 81.14/30.17 & 49.4/53.4 & -           \\ \hline
\end{tabular}
\end{table*}

On clean MNIST, without FM-Defense, the accuracy of the classifier is $99.4 \%$; with FM-Defense, the accuracy is reduced to  $98.3\%$. This small reduction is negligible. 
As illustrated in the table, the defense performance of FM-Defense with maximum defense abilities of MagNet, and Defense-GAN. Table II shows the performances of defense methods on the MNIST and the FMNIST datasets, respectively. 
As shown, the performance of FM-Defense exceeds that of MagNet on all oblivious attacks (DeepFool and CW). FM-Defense also outperforms FB-GAN and Defense-GAN on the FGSM attack.
Overall, FM-Defense shows the best performance against all evaluated attacks. 
These evaluations provide empirical evidence that FM-Defense is effective, easy to conduct and generalizes well to different attacks.
\section{Conclusion}
In this paper, we propose FM-Defense, a one-off and attack-agnostic defense that effectively detects and purifies state-to-the-arts adversarial attacks. 
The assumption of our defense is that the perturbation is lack of transferability. 
FM-Defense applies a combo-VAE for both the adversarial detection and salvageable instance purification. FM-Defense achieves high accuracy against the state-of-the-art attacks, especially for complex images, delivering empirical evidence that our assumptions are likely correct. 
The combo-VAE with two enhanced encoders is used to reconstruct salvageable candidates for purification. 
However, before we discover stronger justification or proof, it is not appropriate to dismiss the possibility that the good performance of FM-Defense is attributed to the state-of-the-art attacks that are not strong enough. 
We hope that our findings would provoke further explorations on designing more powerful attacks even more practical anomaly detection.

In this work, the adversaries are assumed to have no knowledge about the detector and purifier. Even if the attacker knows everything else about the defense, such as network structure, training
set, and training procedure, the randomness derived from cryptography could be applied to guarantee the computational difficulty for the attacker. Specifically, we can generate a great number and large diversity of autoencoders candidates and randomly select one of these autoencoders for each defensive device for every session, every test set, or even every test example. 



\bibliographystyle{IEEEtran}
%

\bibliography{bare_conf}

\end{document}